\documentclass[10pt,twocolumn,letterpaper]{article}

\usepackage{cvpr}              

\usepackage{graphicx}
\usepackage{amsmath}
\usepackage{amssymb}
\usepackage{booktabs}
\usepackage[dvipsnames]{xcolor}
\usepackage[accsupp]{axessibility} 
\usepackage{enumitem,kantlipsum}

\usepackage[pagebackref,breaklinks,colorlinks]{hyperref}

\usepackage[capitalize]{cleveref}
\crefname{section}{Sec.}{Secs.}
\Crefname{section}{Section}{Sections}
\Crefname{table}{Table}{Tables}
\crefname{table}{Tab.}{Tabs.}

\newcommand{\KGnote}[1]{{\color{magenta}{\bf KG: }#1}}
\newcommand\KG[1]{\textcolor{blue}{#1}}
\newcommand\KGnew[1]{\textcolor{red}{#1}}

\newcommand\cc[1]{\textcolor{brown}{#1}}

\newcommand{\RGnote}[1]{{\color{blue}{\bf RG: }#1}}

\newcommand\CA[1]{\textcolor{brown}{#1}}
\newcommand\CAn[1]{\textcolor{black}{#1}} 
\newcommand\CAcr[1]{\textcolor{brown}{#1}}
\newcommand{\CAnote}[1]{{\color{brown}{\bf CA: }#1}}

\newcommand{\PCnote}[1]{{\color{orange}{\bf PC: }#1}}

\renewcommand{\PCnote}[1]{}
\renewcommand{\CAnote}[1]{}
\renewcommand{\RGnote}[1]{}
\renewcommand{\KGnote}[1]{}
\renewcommand\KG[1]{\textcolor{black}{#1}}

\newcommand\KGcr[1]{\textcolor{blue}{#1}}
\renewcommand\cc[1]{\textcolor{black}{#1}}

\renewcommand\CA[1]{\textcolor{black}{#1}}
\renewcommand\CAn[1]{\textcolor{black}{#1}}
\renewcommand\KGnew[1]{\textcolor{black}{#1}}
\renewcommand\CAcr[1]{\textcolor{black}{#1}}
\renewcommand\KGcr[1]{\textcolor{black}{#1}}

\def\cvprPaperID{9896} 
\def\confName{CVPR}
\def\confYear{2022}

\begin{document}

\title{Visual Acoustic Matching}

\author{Changan Chen$^{1,4}$ \hspace{3mm}  Ruohan Gao$^{2}$ \hspace{3mm} Paul Calamia$^{3}$ \hspace{3mm} Kristen Grauman$^{1,4}$\\
$^1$University of Texas at Austin \hspace{1mm} $^2$Stanford University\hspace{1mm}  $^3$Reality Labs at Meta \hspace{1mm} $^4$Meta AI
}
\date{\today}

\maketitle

\begin{abstract}
\KG{We introduce the \emph{visual acoustic matching} task, in which an audio clip is transformed to sound like it was recorded in a target environment. Given an image of the target environment and a waveform for the source audio, the goal is to re-synthesize the audio to match the target room acoustics as suggested by its visible geometry and materials. To address this novel task, we propose a cross-modal transformer model that uses audio-visual attention to inject visual properties into the audio and generate realistic audio output.  In addition, we devise a self-supervised training objective that can learn acoustic matching from in-the-wild Web videos, despite their lack of acoustically mismatched audio. We demonstrate that our approach successfully translates human speech to a variety of real-world environments depicted in images, outperforming both traditional acoustic matching and more heavily supervised baselines.}
\end{abstract}
\section{Introduction}

\KG{The audio we hear is always transformed by the space we are in, as a function of the physical environment's geometry, the materials of surfaces and objects in it, and the locations of sound sources around us.  This means that we perceive the same sound  differently depending on where we hear it.  For example, imagine a person singing a song while standing on the hardwood stage in a spacious auditorium versus in a cozy living room with shaggy carpet.  The underlying song content would be identical, but we would experience it in two very different ways.}

\KG{For this reason, it is important to model room acoustics to deliver a realistic and immersive experience for many applications in augmented reality (AR) and virtual reality (VR).
Hearing sounds with acoustics \textit{inconsistent} with the scene is disruptive for human perception. In AR/VR, when the real space and virtually reproduced space have different acoustic properties, it causes a cognitive mismatch and the ``room divergence effect"  damages the user experience~\cite{Werner21}.}

\begin{figure}[t]
    \centering
    \includegraphics[width=\linewidth]{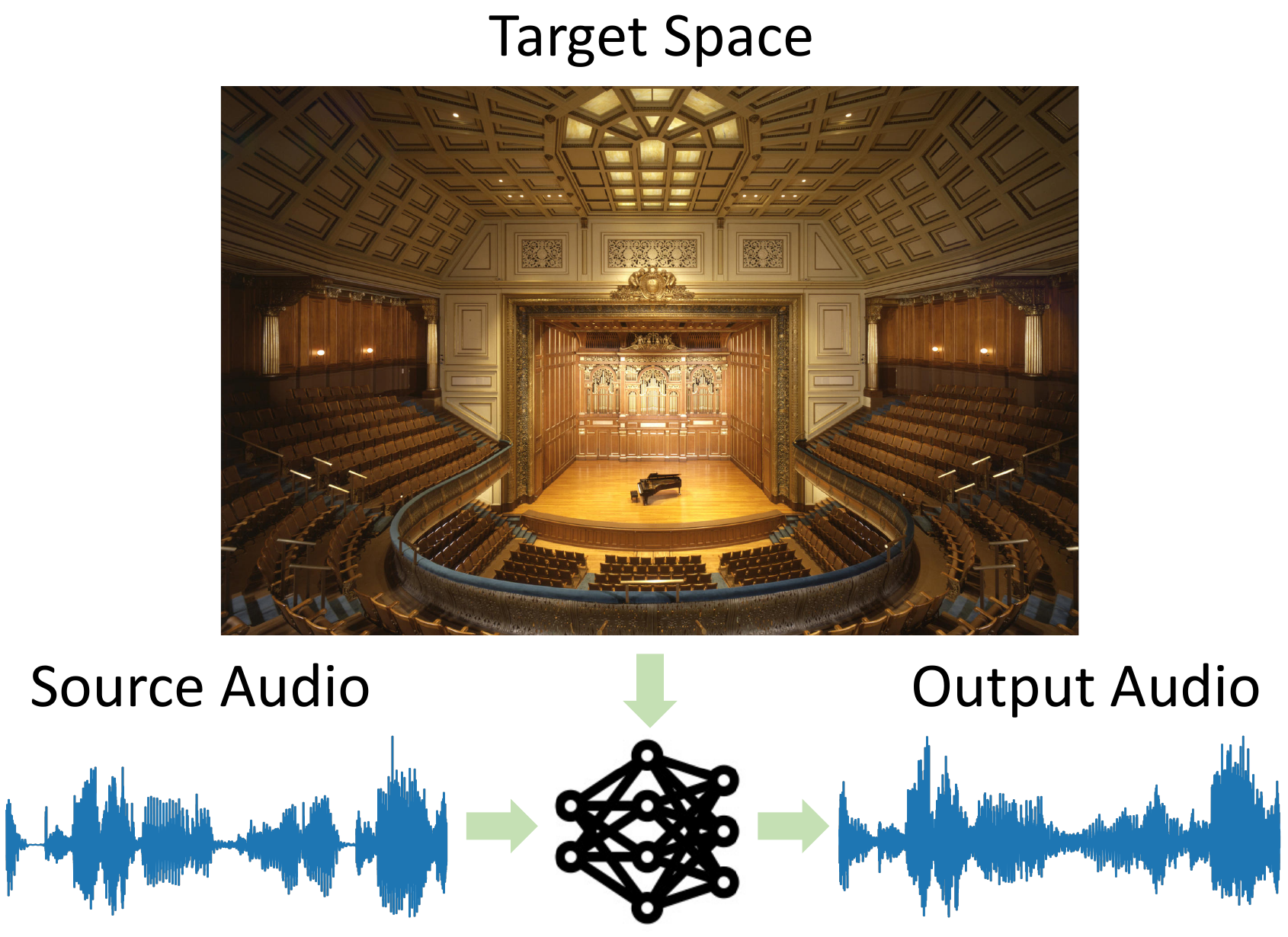}
        \vspace{-0.1in}
    \caption{\KG{Goal of} visual acoustic matching: transform the sound recorded in one space to another space depicted in the \KG{target} visual scene. For example, 
    \KG{given source audio recorded in a studio, re-synthesize that audio}
    to match the room acoustics of 
    a concert hall.
    }
    \vspace{-0.1in}
    \label{fig:concept}
\end{figure}

Creating audio signals that are consistent with an environment has a long history in the audio community. If the geometry (often in the form of a 3D mesh) and material properties of the space are known, simulation techniques can be applied to generate a room impulse response (RIR), a transfer function between the sound source and the microphone that describes how the sound gets transformed by the space. RIRs can then be convolved with an arbitrary source audio signal to generate the audio signals 
received by the microphone~\cite{bilbao2013modeling, cao2016interactive, funkhouser2004beam, savioja2015overview, savioja2020simulation}. In the absence of geometry and material information, the acoustical properties 
can be estimated blindly from audio captured in that room (e.g., reverberant speech), 
then used to auralize a signal~\cite{klein2019real, murgai2017blind,su2020acoustic}. 
\KG{However, both approaches have practical limitations:} the former requires access to the full mesh and material properties of the target space, while the latter gets only limited acoustic information about the target space from the reverberation in the audio sample.
\KG{Neither uses imagery of the target scene to perform acoustic matching.}

We propose a novel task: \emph{visual acoustic matching}.  \KG{Given} an image of the target environment and a source audio clip, \KG{the goal is} to re-synthesize the audio as if it \KGnew{were} recorded in the target environment (see Figure~\ref{fig:concept}). 
The idea is to transform sounds from one space to another space by altering their scene-driven acoustic signatures. 
\KG{Visual acoustic matching has many potential applications}, including smart video editing \KG{where a user can inject sounding objects into new backgrounds}, film dubbing \KG{to make a different actor's voice sound appropriate for the movie scene}, audio enhancement for video conference calls, and audio synthesis for AR/VR \KG{to make users feel immersed in the visual space displayed to them.}

\KG{To address visual acoustic matching, we introduce a cross-modal transformer model together with a novel self-supervised training objective that accommodates in-the-wild Web videos having unknown room acoustics.}

\KG{Our approach accounts for two key challenges: how to faithfully model the complex cross-modal interactions, and how to achieve scalable training data.}  
\KG{Regarding the first challenge, different regions of a room  affect the acoustics in different ways.}
For example, reflective glass leads to longer reverberation in high frequencies while absorptive ceilings reduce the reverberation more quickly. 
Our model provides fine-grained audio-visual reasoning by attending to regions of the image and how they affect the acoustics.
Furthermore, to capture the fine details of reverberation effects---which are typically much smaller in magnitude than the direct signal---we use 1D convolutions to generate time-domain signals directly and apply a multi-resolution generative adversarial audio loss.

\KG{Regarding the second key challenge, one would ideally have \emph{paired} training data consisting of a sound sample not recorded in the target space plus its proper acoustic rendering for the scene shown in the target image, i.e., a source and target audio for each visual scene in the training set.  
However, such a strategy requires either physical access to the pictured environments, or knowledge of their room impulse response functions---either of which severely limits the source of viable training data.  
Meanwhile, though a Web video does exhibit strong correspondence between its visual scene and the scene acoustics, it offers only the audio recorded in the target space. 
Accounting for these tradeoffs, we propose a self-supervised objective  that 
automatically creates acoustically mismatched audio for training with Web videos.  
The key insight is to use dereverberation and acoustic randomization to alter the original audio's acoustics while preserving its content.}

\KG{We demonstrate our approach on challenging real-world sounds and environments, as well as controlled experiments with realistic acoustic simulations in scanned scenes.}  Our quantitative results and subjective evaluations via human studies show that our model generates audio that matches the target environment with high perceptual quality, \KG{outperforming a state-of-the-art model that has heavier supervision requirements~\cite{singh_image2reverb_2021} as well as traditional acoustic matching models.}
\section{Related Work}\label{sec:related_work}

\paragraph{Acoustic matching.} 
The goal of \emph{acoustic matching} is to transform an audio recording made in one environment to 
\KG{sound} as if it were recorded in a target environment. 
The audio community deals with this task with various 
approaches depending on what information about the target environment is accessible. 
If audio recorded in the target environment is provided, blind estimation of two acoustic 
parameters, direct-to-reverberant ratio (DRR), \cc{which describes the energy ratio of direct arrival sound and reﬂected sound,} and reverberation time (RT60), \cc{the time it takes for a sound to decay 60dB,} is sufficient to create simple RIRs that yield plausibly matched audio~\cite{eaton20161ACE, gamper2018blind, klein2019real, mack2020single, murgai2017blind, xiong2018joint}. 
Blind estimation of the room impulse response from reverberant speech has also been explored~\cite{steinmetz20218blind, wager2020dereverberation}.
In music production, acoustic matching is applied to change the reverberation 
to emulate that \KGcr{of} a target space or processing algorithm~\cite{koo2021reverb, sarroff2020blind}. Recent work 
conditions the target-audio generation on a low-dimensional audio embedding~\cite{su2020acoustic}. 
\KG{Unlike any of the above,} 
we \KG{introduce and} tackle the \emph{visual} acoustic matching problem, where the 
target environment \KG{is expressed via an input image}.

\vspace{-0.1in}
\paragraph{Visual understanding of room acoustics.}
The room impulse response (RIR) is the (time-domain) transfer function capturing the room acoustics for arbitrary source stimuli given specific source and receiver/listener positions in an environment. Convolving an RIR with a sound waveform yields the sound of that source in the context of the particular physical space. RIRs are traditionally measured with special equipment in the room itself~\cite{stan2002comparison,Holters_rir} or simulated with sound propagation models ~\cite{image_method,chen_soundspaces_2020,Murphy_waveguide}. Recent work explores estimating an RIR from an input image~\cite{kon_estimation_2019,singh_image2reverb_2021,Majumder2022_fewshot}, \KG{which requires} 
access to paired image and impulse response \KG{training} data. \cc{While video recordings provide a natural source for learning the correspondence between space (captured by the visual stream) and acoustics (captured by the audio stream), they have not been explored in the literature.} 
\KG{We show how to} leverage Web video data for understanding room acoustics in a self-supervised fashion, \KG{obviating the need for expensive paired RIR-image training data.  Our results demonstrate the advantages.}

\vspace{-0.1in}
\paragraph{Audio-visual learning.}
\KG{Recent advances in} multi-modal video understanding 
\KG{enable new forms of} self-supervised cross-modal feature learning from video~\cite{lorenzo-nips2020,korbar-nips2018,morgado-spatial-nips2020}, object localization~\cite{hu-localize-nips2020}, and audio-visual speech enhancement and source separation~\cite{ephrat2018looking,owens2018audio,zhao2019som,Afouras18,Afouras20audio-visual-objects,Michelsanti19,Sadeghi20,zhou19,Hou17,chen_dereverb21}. \KG{Work in embodied AI explores} 
acoustic simulations with real visual scans to study audio-visual 
\KG{navigation tasks}~\cite{chen_soundspaces_2020,gan2019look,dean-curious-nips2020,chen_savi_2021,YinfengICLR2022saavn,chen_waypoints_2021}, where an agent \KG{moves intelligently} based on the visual and auditory observations. 
However, no prior work investigates \KG{the visual acoustic matching task as we propose}.  

\vspace{-0.1in}
\paragraph{Multimodal fusion.}
One standard solution for audio-visual feature fusion is to represent audio as spectrograms, a \KGnew{matrix} representation of the spectrum of frequencies of a signal as it varies with time, process them with a CNN, and concatenate with visual features from another CNN\cite{owens2018audio,25d-visual-sound,ephrat2018looking,gao_visualvoice_2021,chen_soundspaces_2020}.  This fusion strategy is limited by using one global feature to represent the scene and thus \KG{supports only} coarse-grained reasoning.
The transformer~\cite{vaswani2017attention} has proven to be a power tool in vision~\cite{vit,girdhar2021anticipative}. Its self-attention operation provides a natural mechanism to fuse high-dimensional signals of different sensory modalities, and it has been used in various tasks such as action recognition~\cite{bertasius2021space}, self-supervised learning~\cite{lorenzo-nips2020,akbari2021vatt,patrick2021space}, and language modeling~\cite{conformer}.
Audio-visual attention~\cite{tsai2019MULT,Truong2021ICCV,Lin2020Audiovisual} has been recently studied to capture the correlation between visual features and audio features.
We use cross-modal attention for learning how different regions of the image contribute to reverberation. \KG{We show that} compared with the \KG{conventional} concatenation-based fusion, the proposed model predicts acoustics from images more accurately.

\section{The Visual Acoustic Matching Task}

We introduce a novel task, \textit{visual acoustic matching}. In this task, an audio recording $A_S$ recorded in space $S$ and an image $I_T$ of a different target space $T$ are provided as input. The goal is to predict $A_T$, which has the same audio content as $A_S$ but sounds as if it \KGnew{were} recorded in space $T$ \KG{with a microphone co-located with $I_T$'s camera}.  \KG{Our goal is thus to learn a function $f$ such that $f(A_S,I_T) = A_T$.}
The microphone co-location is important because acoustic properties vary as the listener location changes; inconsistent camera locations would lead to a perceived mismatch between the visuals and acoustics. 
The space $S$ can have arbitrary acoustic characteristics, from an anechoic recording studio to a concert hall with significant reverberation.  We assume there is one sounding object, \KG{leaving the handling of  background sounds or interference as future work.} 

Importantly, our task formulation does \emph{not} \KG{assume} 
access to the impulse response, nor does it require the input \KG{audio} 
to be anechoic.
In comparison, the Image2Reverb~\cite{singh_image2reverb_2021} task requires access to both the impulse response and clean input audio, and 
\KG{does not account for the co-location of the camera and microphone.}

\section{Datasets}\label{sec:datasets}

\begin{figure}
    \centering
    \includegraphics[width=0.8\linewidth]{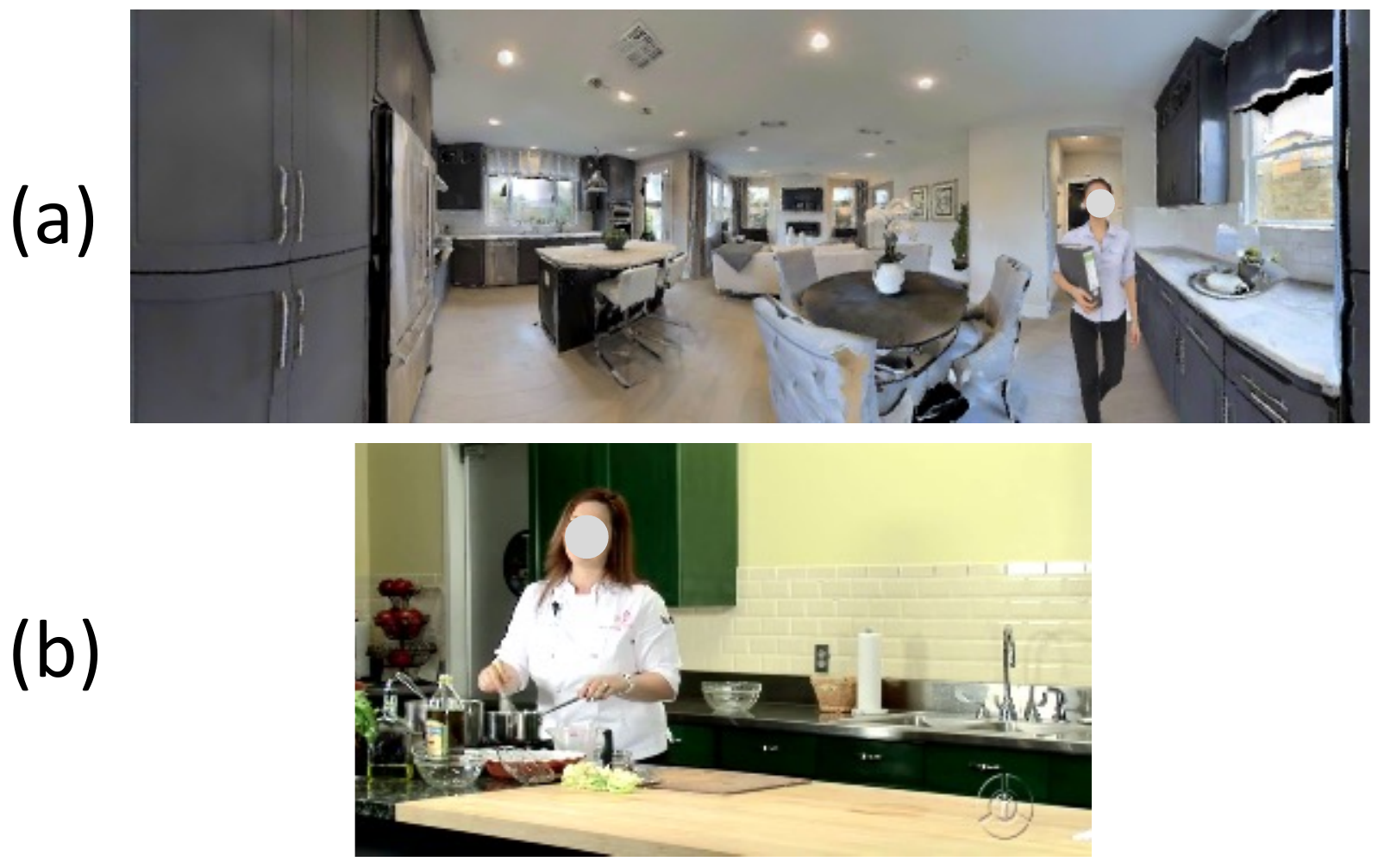}
    \vspace{-0.05in}
    \caption{\KGnew{Example} images in (a) SoundSpaces and (b) AVSpeech. 
    }
    \vspace{-0.2in}
    \label{fig:data}
\end{figure}

\KG{We consider two datasets: simulated audio in scanned real-world environments (Sec.~\ref{sec:syntheticdata}), and in-the-wild Web videos with their recorded audio (Sec.~\ref{sec:realdata}).  The former has the advantage of clean paired training data for $A_T$ and $A_S$ as well as precise ground truth for evaluating the output audio, but necessarily has a realism gap. The latter has the advantage of total realism, but makes quantitative evaluation more complex.}

\KG{For both, we focus on human speech in indoor settings given its relevance to many of the applications cited above, and due to the fact that human listeners have strong prior knowledge about how reverberation should affect speech.  However, our model design is not specific to speech.} 
\CAcr{See Supp. for its applicability on non-speech sounds.}

\begin{figure*}[t]
    \centering
    \includegraphics[width=\linewidth]{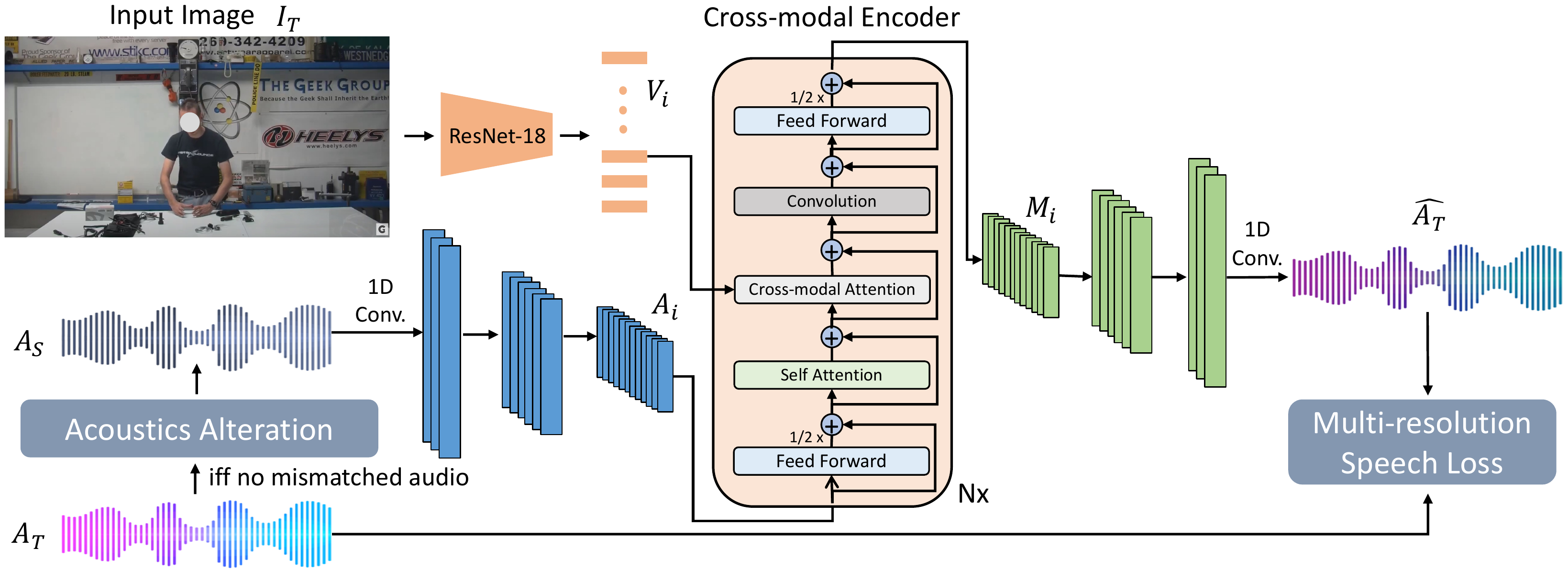}
    \vspace{-0.15in}
    \caption{\KG{AViTAR} model illustration. We extract visual feature sequence $V_i$ from input image $I_T$ with a ResNet-18~\cite{resnet18}, and audio feature sequence $A_i$ from input audio $A_S$ with 1D convolutions. $V_i$ and $A_i$ are passed into cross-modal encoders for cross-modal reasoning. The output feature sequence $M_i$ is processed and upsampled with 1D convolutions to recover the \KG{output of the} same temporal length. Finally, we use a multi-resolution speech GAN loss to guide the audio synthesis to be high fidelity. The acoustics alteration process is applied to the target audio \KG{during training} if and only if there is no mismatched audio, e.g., on the Acoustic AVSpeech dataset. 
    }
    \vspace{-0.1in}
    \label{fig:model}
\end{figure*}

\subsection{
\KG{SoundSpaces-Speech Dataset}}\label{sec:syntheticdata}
With 
\KG{the SoundSpaces platform}~\cite{chen_soundspaces_2020}, acoustics can be accurately simulated based on 3D scans of real-world environments~\cite{Matterport3D,straub2019replica,xiazamirhe2018gibsonenv}.
\KG{This allows highly realistic rendering of arbitrary camera views and arbitrary microphone placements for waveforms of the user's choosing, accounting for all major real-world
audio factors: direct sounds, early specular/diffuse
reflections, reverberation, binaural spatialization, and 
effects from materials and air absorption.}

\KG{We adopt a SoundSpaces-Speech dataset created in~\cite{chen_dereverb21}}
consisting of paired clean \KG{(anechoic)} and reverberant \KG{audio samples together with camera views}.\KG{\footnote{Note that~\cite{chen_dereverb21} uses the data for dereverberation, not acoustic matching.}} 
The RIRs for 82 Matterport3D~\cite{Matterport3D} environments 
are convolved with non-overlapping speech clips from LibriSpeech~\cite{librispeech}. 
A 3D humanoid of the same gender as the real speaker is inserted at the speaker location and panorama RGB-D images are rendered at the listener location.  \KG{See Figure~\ref{fig:data}a.} 
\KG{Excluding those samples}
where the speaker is very distant or out-of-view (for which the visual input does not capture the geometry of the source location), 
there are 28,853/1,441/1,489 samples for \KG{the train/val/test splits}.

\subsection{Acoustic AVSpeech \KG{Web Videos}}\label{sec:realdata}
Web videos offer rich and natural supervision for the association between  visuals and acoustics.
\KG{We adopt a subset of the AVSpeech~\cite{ephrat2018looking} dataset, which contains 3-10 second YouTube clips from 290k videos of single (visible) human speakers without interfering background noises.} 
\KG{We automatically filter the full dataset down to those clips likely to meet our problem formulation criteria: 1) microphone and camera should be co-located and at a position different than the sound source (so that the audio contains not only the source speech but also the reverberation caused by the environment), and 2) audio recording should be reverberant (so that the  physical space has influenced the audio). Cameras in this dataset are typically static, and thus we use single frames and their corresponding audio for this task.
See Supp.~for details.}
This yields 113k/3k/3k video clips for train/val/test splits.  We refer to this \KG{filtered} 
dataset as Acoustic AVSpeech.  \KG{See Figure~\ref{fig:data}b.}

\section{Approach}\label{sec:approach}
We present the \textbf{A}udio-\textbf{Vi}sual \textbf{T}ransformer for \textbf{A}udio Gene\textbf{r}ation model (\textbf{AViTAR}) (Figure~\ref{fig:model}).
AViTAR learns to perform cross-modal attention based on \KG{sequences of} convolutional features of audio and images and then synthesizes the desired waveform \KGnew{$\hat{A_T}$}. 
\KG{We first define the audio-visual features (Sec.~\ref{sec:features}) and their cross-modal attention  (Sec.~\ref{sec:crossmodal}), followed by our approach to waveform generation (Sec.~\ref{sec:waveform}).  Finally, we present our acoustics alteration idea to enable learning from in-the-wild video (Sec.~\ref{sec:randomize}).}

\subsection{Audio-Visual Feature Sequence Generation}\label{sec:features}
To apply cross-modal attention, we first need to generate sequences of audio and visual features, where each element in the sequence represents features of a part of the input space. For visual sequence generation \KG{from image $I_T$}, we use ResNet18~\cite{resnet18}
and flatten the last feature map before the pooling layer, yielding the visual feature sequence $V_i$. 

For audio feature sequence generation \KG{from source audio $A_S$}, we generate audio features $A_i$ from the waveform directly with stacked 1D convolutions. We first use one 1D conv layer to embed the input waveform into a latent space. We then apply a sequence of strided 1D convolutions, each doubling the channel size while downsampling the input sequence. The output audio features are a sequence of vectors of size $S$, with length downsampled $D$ 
times from the input. Weight normalization is applied to 1D conv layers. We 
rather than STFT \KG{spectrograms so that the} audio features are not limited to one resolution and can be optimized end-to-end to learn the most important features for the visual acoustic matching task.

\subsection{Cross-Modal Encoder}\label{sec:crossmodal}
\KG{Prior work} often models audio-visual \KG{inputs}
in a simplistic manner by representing the image feature with one single vector and concatenating it with the audio feature~\cite{owens2018audio,25d-visual-sound,ephrat2018looking,zhao2019som,gao_visualvoice_2021,chen_dereverb21,chen_soundspaces_2020}. 
\KG{However,} 
for visual acoustic matching, it is important to reason how different regions of the space contribute to the acoustics differently. For example, a highly reflective glass door leads to longer reverberation time for high frequencies, while absorptive ceilings diminish that quickly.
\KG{Thus,} we propose to attend to image regions to reason how different image patches contribute to the acoustics, leveraging recent advances on the transformer architecture~\cite{vaswani2017attention,vit,conformer}.  

For cross-modal attention, we first adopt the conformer variant~\cite{conformer} of encoder blocks, which adds one convolution layer inside the block for modeling local interaction for speech features. Based on this block, we insert one cross-modal attention layer \KG{$\mathcal{A}_{cm}$} after the first feed-forward layer, described as follows:
\begin{equation}
    \mathcal{A}_{cm}(A_i, V_i) = \textrm{softmax} (\frac{A_i V_i^T}{\sqrt{S}})V_i,
\end{equation}
where  the attention scores between the two sequences of features $A_i$ and $V_i$ are first calculated by dot-product,  then normalized by softmax, scaled by $\frac{1}{\sqrt{S}}$, and finally used to weight the visual features $V_i$.
This cross-modal attention allows the model to attend to different image region features and reason about how they affect the reverberation. 
Absolute positional encoding is added to the visual encoding. After passing $V_i$ and $A_i$ through $N$ encoder blocks, we obtain the fused audio-visual feature sequence $M_i$, which has the same length as $A_i$.

\subsection{Waveform Generation and Loss}\label{sec:waveform}

\KG{Recent audio-visual work generates audio outputs by inferring spectrograms then using ISTFT reconstruction to obtain a waveform (e.g., \cite{Yamamoto2020wavegan,gao_visualvoice_2021,ephrat2018looking,25d-visual-sound,zhao2019som,Zhao_2018_ECCV}).  While sensible for source separation, where the}
target signal is a subset of the source signal,
ratio mask prediction 
\KG{is inadequate for our task,}
because reverberation might occupy periods of silence in the input audio and the ratio will be unbounded (\KG{as we verify in results)}.
\KG{Futhermore,} generating audio based on spectrograms is limiting 
because 1) predicting the coherent phase component remains challenging~\cite{ai2019phase,Choi2019phase}, and 2) the spectrogram has one fixed resolution (one FFT size, hop length, and window size).

\KG{Instead, we aim to synthesize}
 time-domain signals directly, skipping the intermediate spectrogram generation step and allowing more flexibility for what losses can be imposed, inspired by recent advances on time-domain speech synthesis~\cite{Oord2016wavnet,Prenger2018waveglow,kong2020hifigan,kumar2019melgan}.
Specifically, with the fused audio-visual feature sequence $M_i$, we apply a sequence of transposed strided 1D convolutions, each halving the channel size while upsampling the input sequence, which is exactly the reverse operation of the audio encoding. 
Altogether, we upsample the audio sequence $D$ times and obtain a waveform of the same length as the input.

\KG{Next we incorporate a multi-resolution generative loss.}
We found directly minimizing a Euclidean distance based loss \KG{between the target ground truth audio $A_T$ and the inferred audio $\hat{A_T}$} leads to distortion in the generated audio on this task \KG{(cf.~Figure~\ref{fig:qual} \KGnew{and Tab.~\ref{tab:ablations}}).} Therefore, to let the model learn how to reverberate the input speech properly, \KG{we employ a generative adversarial loss }\CA{where a set of discriminators operating at different resolutions are trained to identify reverberation patterns and guide the generated audio to sound like real examples.}
Specifically, we apply an adversarial loss~\cite{kong2020hifigan} \KG{comprised of the generator and discriminator losses}: 
\begin{equation}
\begin{split}
    \mathcal{L}_G &= \sum_{k=1}^{K}(\mathcal{L}_{Adv}(G;D_k) + \lambda_{1} \mathcal{L}_{FM}(G;D_k)) + \lambda_{2}\mathcal{L}_{Mel}(G), \\
    \mathcal{L}_D &= \sum_{k=1}^{K}\mathcal{L}_{Adv}(D_k;G),\nonumber
\end{split}
\vspace{-0.15in}
\end{equation}
\KG{where each} $D_k$ is a sub-discriminator that operates at one of $K$ different scales and periods for distinguishing the fake and real examples. $\mathcal{L}_{Adv}$ is the LS-GAN~\cite{mao2016lsgan} training objective, which trains the generator to fake the discriminator and trains the discriminator to distinguish real examples from fake ones.
For the generator $G$, a feature matching loss~\cite{kumar2019melgan}  $\mathcal{L}_{FM}$ is used, which is a learned similarity metric measured by the difference in features of the discriminator between a ground truth sample and a generated sample. An additional mel-spectrogram loss $\mathcal{L}_{Mel}$ is imposed on the generator for improving the training efficiency and fidelity of the generated audio. $\lambda_{1}$ and $\lambda_{2}$ are two weighting factors for these two losses. The generator loss $\mathcal{L}_G$ and discriminator loss $\mathcal{L}_D$ are trained alternatively competing against each other. For more details, refer to~\cite{kong2020hifigan}.

\begin{figure}
    \centering
    \includegraphics[width=\linewidth]{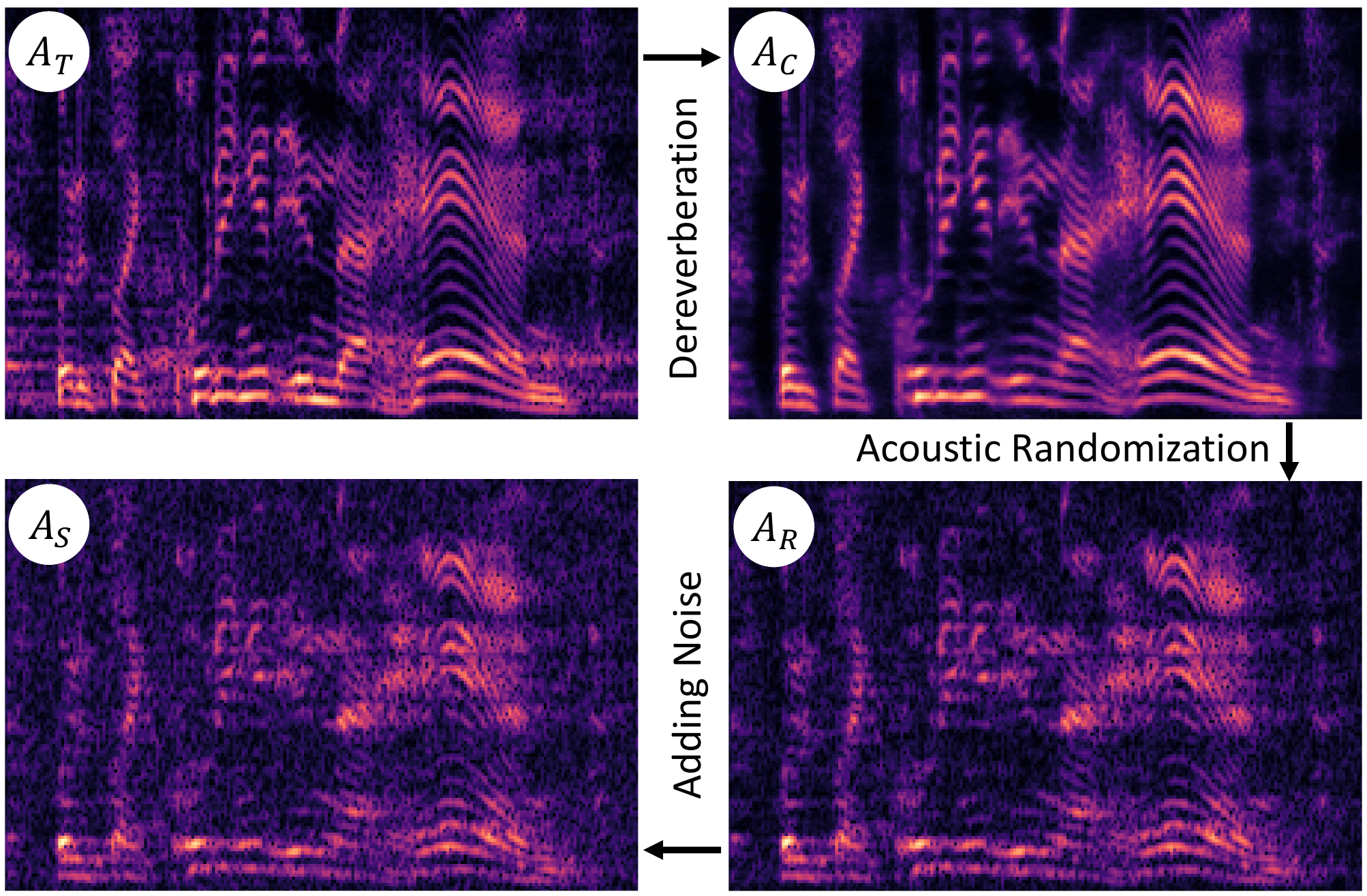}
    \vspace{-0.2in}
    \caption{Acoustics alteration process. Spectrograms of the resulting audio after each step are shown. \CA{We first dereverberate the target audio $A_T$ to obtain cleaner audio $A_C$, randomize its acoustics by applying an impulse response of another environment to obtain $A_R$, and finally, add Gaussian noise to $A_R$ to create $A_S$. Notice how the spectral pattern changes in this process.}  
    }
    \vspace{-0.1in}
    \label{fig:alteration}
\end{figure}

\subsection{Acoustics Alteration for Self-Supervision} 
\label{sec:randomize}

\KG{The training paradigm differs in one important way depending on the source of training data (cf.~Sec.~\ref{sec:datasets}).  For the simulated SoundSpaces data, we have access to an anechoic audio sample $A_S$ as well as the ground truth reverberated sample $A_T$ as it should be rendered in the target environment for a camera seeing view $I_T$.  This means we can train to (implicitly) discover the mapping that takes the target image to an RIR which, when convolved with $A_S$, yields $A_T$. }

\KG{For the in-the-wild video data (AVSpeech), however, we have only $A_T$ and $I_T$ to train, i.e., we only observe sounds that \emph{do} match their respective views.}
Thus, to leverage unannotated Web video, we need to create an audio clip \KG{that preserves the target audio content but} has \emph{mismatched} acoustics. Figure~\ref{fig:alteration} illustrates the steps for this process.  
First we strip away the original acoustics of the target environment by performing dereverberation \KG{on the audio $A_T$ alone} with the pretrained model from \cite{chen_dereverb21}. 
Since dereverberation is imperfect, there is residual acoustic information in the dereverberated output $A_C$, meaning that the resulting ``clean" audio is still predictive of the \KG{target} environment.

Thus, we subsequently randomize the acoustics by convolving that audio with an impulse response of another environment, yielding $A_R$; that IR is randomly chosen from the corresponding train/val/test split of SoundSpaces-Speech.
The idea is to transform the \KG{semi-clean intermediate} sound into another space to create more acoustic confusion, \KG{thereby forcing the model to learn from the target image}.  Finally, to further suppress the residual acoustics from the training environment, we add Gaussian noise with SNR randomly sampled from 2-10 dB to $A_R$ and obtain \KG{the training source audio} $A_S$. \CA{See more details about how each step alters the acoustics in Supp.}
In short, with this strategy, we are able to leverage readily available Web videos for our proposed task, \KG{despite its lack of ground truth paired audio.}

\begin{table*}[t]
\centering
\setlength{\tabcolsep}{5pt}
\begin{tabular}{ c|c|c|c|c|c|c|c|c|c|c} 
\toprule
    & \multicolumn{6}{c|}{\textit{\KG{SoundSpaces-Speech}}} & \multicolumn{4}{c}{\textit{Acoustic AVSpeech}} \\
    & \multicolumn{3}{c|}{\textit{Seen}} & \multicolumn{3}{c|}{\textit{Unseen}} & \multicolumn{2}{c}{\textit{Seen}} & \multicolumn{2}{c}{\textit{Unseen}} \\
            & STFT & RTE (s) & MOSE & STFT & RTE (s) & MOSE  & RTE (s) & MOSE & RTE (s) & MOSE \\
\midrule
 Input audio           & 1.192 & 0.331 & 0.617 & 1.206 & 0.356 & 0.611 & 0.387	& 0.658 & 0.392 & 0.634 \\ 
 Blind Reverberator~\cite{Vesa2016}   & 1.338 & 0.044 & 0.312 & - & - & - &  -	& - & - & - \\ 
 Image2Reverb~\cite{singh_image2reverb_2021}  & 2.538 & 0.293 & 0.508  & 2.318 & 0.317 & 0.518 & - & - & - & - \\ 
 AV U-Net~\cite{25d-visual-sound}  & \textbf{0.638} & 0.095 & 0.353  & \textbf{0.658} & 0.118 & 0.367 & 0.156 & 0.570 & 0.188 & 0.540 \\ 
\midrule
 AViTAR w/o visual     & 0.862 & 0.140 & 0.217 & 0.902 & 0.186 & 0.236 & 0.194 & 0.504 & 0.207 & 0.478 \\ 
 AViTAR                  & 0.665 & \textbf{0.034} & \textbf{0.161} & 0.822 & \textbf{0.062} & \textbf{0.195}  & \textbf{0.144} & \textbf{0.481} & \textbf{0.183} & \textbf{0.453} \\ 
\bottomrule
\end{tabular}
\vspace{-0.05in}
\caption{Results on the SoundSpaces-Speech and Acoustic AVSpeech datasets \KG{for Seen and Unseen environments}.  All input audio at test time is novel (unheard during training). 
\KG{Note that the STFT metric is applicable only for SoundSpaces, where we can access the ground truth $A_T$'s spectrogram.} \CA{For all metrics, lower values are better.} 
Standard errors for STFT, RTE and MOSE are all less than 0.04, 0.013s and 0.01 on SoundSpaces-Speech. Standard errors for RTE and MOSE are all less than 0.005s and 0.01 on Acoustic AVSpeech.
}
\vspace{-0.1in}
\label{tab:main}
\end{table*}
\section{Experiment} \label{sec:experiment}
We validate our model on two datasets using comprehensive metrics and baselines. Implementation and training details can be found in Supp.

\vspace{-0.15in}
\paragraph{Evaluation metrics. }  
We measure the quality of the generated audio from three aspects: 
1) the closeness to the ground truth (if ground truth audio is available), \KG{as measured by \textbf{STFT Distance}, i.e., the MSE between the generated and true target audio's magnitude spectrograms;}
2) the correctness of the room acoustics, \KG{as measured by the \textbf{RT60 Error (RTE)} between the true and inferred $A_T$'s RT60 values.  RT60 indicates the reverberation time in seconds for the audio signal to decay by 60 dB, a standard metric to characterize room acoustics. We estimate the RT60 directly from magnitude spectrograms of the output audio, using a model trained with disjoint SoundSpaces data (see Supp.), since impulse responses are not available for the target environments;} 
and 
3) the speech quality \KG{preserved in} the synthesized speech, 
\KG{measured by the \textbf{Mean Opinion Score Error (MOSE)}, which is the difference in speech quality between the true target audio and generated audio, as assessed by a deep learning based objective model MOSNet~\cite{mosnet}.\footnote{\KG{By taking the difference with the true target audio's MOS score (rather than simply the output's score), we account for the fact that properly reverberated speech need not have high speech quality. 
}}}

\KG{Both the RTE and MOSE} metrics are content-invariant and thus useful for evaluation when only audio with correct acoustics and mismatched content is available as ground truth, i.e., Web videos.

In addition, we conduct user studies 
to evaluate whether a given audio is perceived as matching the room acoustics of the reference image. 

\vspace*{-0.15in}
\paragraph{Seen and unseen environments.}
\KG{On both datasets, we evaluate by pairing the source audio $A_S$ with a target image $I_T$ coming from either the training set (\textit{Seen}) or test set (\textit{Unseen}).  The audio is always unobserved in training.
The Seen case is useful to match the audio to scenes where we have video recordings \KG{(e.g., the film dubbing case)}. The Unseen case is important for injecting room acoustics depicted in novel images \KG{(e.g., to match sounds for a random Web photo being used as a Zoom call background)}.}

\vspace{-0.1in}
\paragraph{Baselines. }
We consider the following baselines:
\vspace{-0.05in}
\begin{enumerate}[wide, labelwidth=!, labelindent=0pt]
\item \KG{\textbf{Input audio}.  This is the naive baseline that does nothing, simply returning the input $A_S$ as output.}
\vspace{-0.05in}
\item \CAn{\textbf{Blind Reverberator}.  This is a traditional acoustic matching approach~\cite{Vesa2016} using audio recorded in the target space $T$ as reference with content different from $A_T$. It first 
\KG{estimates} RT60 and DRR 
from the reference audio (estimators are trained using simulated IRs), and then 
synthesizes the \KG{target} IR by shaping an exponentially decaying white noise \KG{based on those two parameters}. 
\KG{Unlike our model, this method requires reference audio at test time and IRs at training time.  It is therefore inapplicable for the Unseen case (no reference audio) and AVSpeech (no training IRs).} 
}
\vspace{-0.15in}
\item \textbf{Image2Reverb~\cite{singh_image2reverb_2021}}. This is a recent approach that trains an IR predictor from images, \KG{then convolves} the predicted IRs with $A_S$
to obtain the target audio. This model requires access to the IR \KG{during training} and thus is not applicable to the Acoustic AVSpeech dataset.  We use the authors' code and convert the \KG{SoundSpaces-Speech} data into the format of their dataset (see Supp.). We replace their depth prediction model with the ground truth depth image, \KG{to improve this baseline's performance.}
\vspace{-0.05in}
\item  \textbf{AV U-Net~\cite{25d-visual-sound}}. This is an audio-visual model originally proposed for visually guided spatial sound generation based on a U-Net network for processing audio spectrograms. We adapt it \KG{for visual acoustic matching} by removing the ratio mask prediction (which we find does not work well).  Instead, we feed in a magnitude spectrogram, predict the target magnitude spectrograms, and generate the time-domain signals with Griffin Lim~\cite{griffin}.  \KG{This baseline helps isolate the impact of our proposed cross-modal attention architecture compared to the common U-Net approach~\cite{25d-visual-sound,owens2018audio,gao_visualvoice_2021,Choi2019phase,Zhao_2018_ECCV}.}
\vspace{-0.05in}
\item \textbf{AViTAR w/o visual}. This model is solely audio-based and is the same as our proposed model except that it does not have visual inputs \KGcr{or} the cross-modal attention layer.  
\end{enumerate}

\subsection{Results on \KG{SoundSpaces-Speech}}
For the SoundSpaces data, we have access to clean anechoic speech, which we use as the input $A_S$.
The simulations offer a clean testbed for this task, showing the potential of each model when it is noise-free and the visuals reveal the full geometry \KG{via the panoramic RGB-D images.}   

Table~\ref{tab:main} \KG{(left)} shows the results. 
As expected, the clean input audio baseline 
\KG{does poorly because it does not account for the target environment}.  
Our AViTAR model has the lowest RT60 error and MOS error,  
indicating that it best predicts the correct acoustics from images, injects them into the speech, and synthesizes high-quality audio. 
The AV U-Net baseline has slightly lower STFT distance than ours, likely because its training objective is to minimize STFT distance. However it has higher \KG{perceptual errors} (RTE and MOSE). 
Image2Reverb's~\cite{singh_image2reverb_2021} high 
errors reveal the difficulty of our task and data, and its inapplicability to AVSpeech highlights our model's self-supervised training advantage.
\CAn{Despite having the estimated RT60 as input (and thus having low RT60 error), Blind Reverberator's STFT  and MOS errors are much higher than AViTAR's,
showing that images are a promising way to characterize room acoustics beyond the traditional RT60.  Plus, its inapplicability for the other scenarios highlights fundamental advantages of AViTAR.
}
Without access to visual information (``w/o visual"), AViTAR can only learn to add an average amount of reverberation to the input audio; 
this confirms that our model successfully learns the acoustics from the visual scene. \cc{Although this variant has higher RT60 error than AV U-Net, its MOS error is lower because the audio quality is better.} See Supp.~video for examples.

\vspace{-0.15in}
\paragraph{Ablations.} Table~\ref{tab:ablations} shows results for ablations on unseen images. For the model architecture, to understand if attending to different image regions with cross-modal attention is helpful, we train the full model with the length of visual feature sequence reduced to one by mean pooling the final ResNet feature map (``w/ pooled visual feature"). This model underperforms the full model on both STFT and RT60 metrics, showing that the audio-visual attention leads to a better visual understanding of room acoustics. 
\KG{Next} we ablate the generative loss and replace it with the non-generative multi-resolution STFT loss~\cite{kumar2019melgan} (``w/o generative loss"), which slightly improves the STFT error but leads to a large drop on the acoustics recovery and speech quality. Despite being multi-resolution, without learnable discriminators to learn to model those fine reverberation details, the audio quality gets worse. 
\CAcr{See Supp.~for GAN loss ablations.}

The synthetic dataset provides access to meta information useful to evaluate whether and how much AViTAR reasons about different visual properties. The location of the sound source matters for acoustics because it directly influences acoustic characteristics like the direct-to-reverberant ratio (DRR). 
When we remove the \KG{3D humanoid from the scene} (``w/o human") in all test images, 
\KG{all error metrics} increase, which indicates that our model reasons \KGnew{about} the location of the sound source \KG{in the image} for accurate acoustic matching. To understand if the model learns meaningful information from the visuals, we replace the target image with a random image (``w/ random image"); this significantly harms our model's performance.

\begin{table}[]
\centering
\begin{tabular}{ c|c|c|c } 
 \toprule
       AViTAR                   & STFT & RTE (s)  & MOSE\\
\midrule
 Full model                     & 0.822 & \textbf{0.062} & 0.195 \\ 
 w/ pooled visual feature          & 0.850 & 0.067 & \textbf{0.193} \\
 w/o generative loss            & \textbf{0.777} & 0.081 & 0.314 \\
 w/o human                      & 0.884 & 0.139 & 0.218 \\ 
 w/ random image                & 0.940 & 0.236 & 0.250 \\ 
 \bottomrule
\end{tabular}
\vspace{-0.05in}
\caption{Ablations on model design and data.  
}
\vspace{-0.25in}
\label{tab:ablations}
\end{table}

\subsection{Results on Acoustic AVSpeech}
\KG{Next, \KGnew{we train} our model on the in-the-wild AVSpeech videos, and test it on novel} clean speech clips from LibriSpeech~\cite{librispeech} ($A_S$) paired with target images ($I_T$)
from AVSpeech.  Here we do not have ground truth for the target speech, so we evaluate 
with RTE and MOSE.

\begin{figure*}[t]
    \centering
    \includegraphics[width=\linewidth]{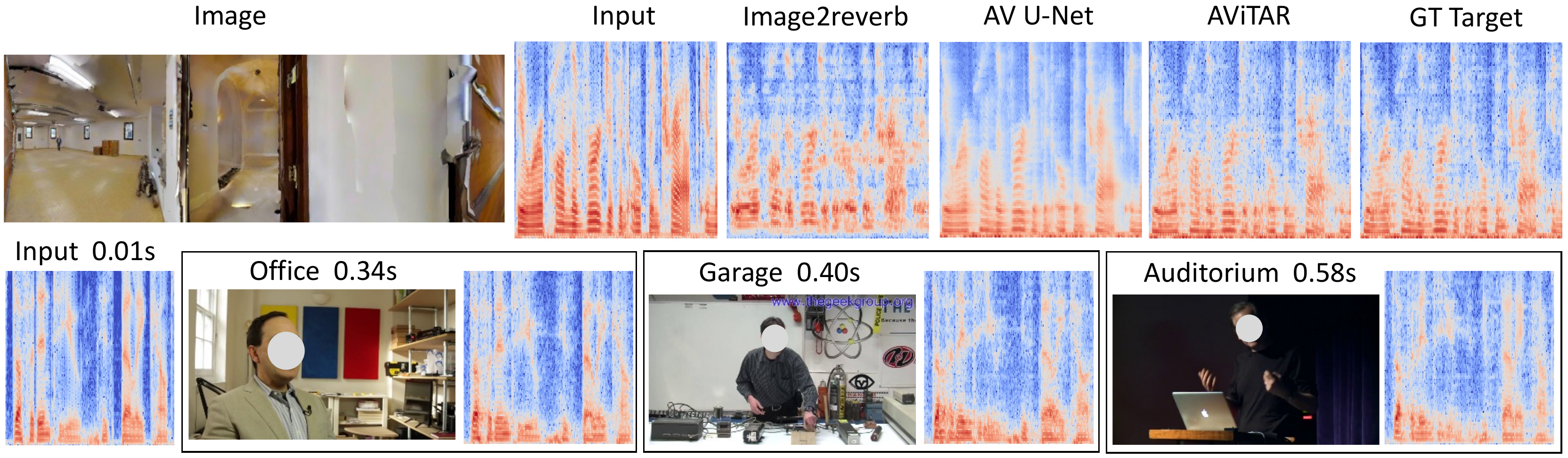}
    \vspace{-0.25in}
    \caption{Qualitative predicted audio. For all audio clips, we compute the magnitude spectrogram, convert the magnitude to dB, and plot the spectrogram with x-axis spanning from 0 to 1.28 s (left to right) and y-axis from 0 to 3000 Hz (bottom to top). Row 1: \KG{SoundSpaces-Speech} example where the target space is a large empty room with a lot of reverberation. Our model predicts the audio  closest to the target clip. \CA{AV U-Net's spectrogram is too smoothed compared to ours and misses some fine reverb details, which leads to perceptual distortion.} Row 2: examples on Acoustic AVSpeech (unseen images). We feed one clean audio clip to match three different scenarios (office, garage, auditorium).  From left to right, the audio spectrogram becomes more reverberant as phoneme patterns get extended and blurred on the temporal axis (est.~RT60 times shown). \KG{NB: AViTAR processes waveforms, not spectrograms; here they are for visualization.}
    }
    \vspace{-0.2in}
    \label{fig:qual}
\end{figure*}

\KG{Table~\ref{tab:main} (right) shows the results.} Our proposed AViTAR model \KG{achieves the lowest RT60 error compared to all baselines.}  This shows our model trained in its self-supervised fashion successfully generalizes to novel images and novel audio, \KG{and demonstrates we can do acoustic matching even for non-anechoic inputs}. AViTAR's MOS error is also the lowest compared to all baselines, showing that it is able to synthesize high-fidelity audio while injecting the proper amount of reverberation into the speech.
\KG{The absolute errors on AVSpeech are higher than on SoundSpaces, which makes sense because the YouTube imagery is more variable, and it has a narrower field of view and no depth, making the geometry and materials of the scene only partly visible.} 
\CAcr{See Supp. for sim2real generalization.}

\begin{table}[]
    \centering
    \begin{tabular}{c|c|c}
    \toprule
    Acoustics Alteration                  & Seen & Unseen \\
    \midrule
        Dereverb. + Randomization + Noise & \textbf{0.144} & \textbf{0.183} \\
        Dereverb. + Randomization         & 0.178 & 0.197 \\
        Dereverb. + Noise                 & 0.170 & 0.208 \\
        Dereverb.                         & 0.230 & 0.250 \\
        $A_T$ + Randomization + Noise     & 0.236 & 0.249 \\
    \bottomrule
    \end{tabular}
    \vspace{-0.1in}
    \caption{Ablations on acoustics alteration. RTE is reported.}
    \vspace{-0.2in}
    \label{tab:randomization_ablations}
\end{table}

\vspace{-0.15in}
\paragraph{Ablations on acoustic alteration.} Table~\ref{tab:randomization_ablations} shows ablations on the proposed acoustics alteration strategy.  In short,  all three steps are necessary to create  an acoustic mismatch with the image, thereby forcing the model to recover the correct acoustics based on the image and allowing better generalization to novel sounds.  \KG{See Supp.~for details.}

\vspace{-0.15in}
\paragraph{User study.} \KG{To supplement the }quantitative metrics and directly 
capture the perceptual quality of the generated samples, we next conduct a user study.  
We show participants the image of the target environment \KG{$I_T$}, the accompanying \KG{ground truth} audio clip \KG{$A_T$} as reference, and paired audio clips \KG{$\hat{A_T}$} generated by AViTAR  and each baseline. We ask participants to select the clip that most sounds as if it were recorded \KGnew{in} the target environment and best matches the reverberation in the given clip. We select 30 reverberant examples \KG{from SoundSpaces-Speech and AVSpeech} and ask 30 participants to complete the assignment on MTurk. 

\begin{table}[t]
    \centering
    \begin{tabular}{c|c|c}
    \toprule
              & SoundSpaces &  AVSpeech \\
    \midrule
        Input Speech                                        & 42.1\% / \textbf{57.9}\%  & 40.1\%  / \textbf{59.9}\% \\
        Image2Reverb~\cite{singh_image2reverb_2021}         & 25.9\% / \textbf{74.1}\%  & - / - \\
        AV U-Net~\cite{25d-visual-sound}                    & 29.8\% / \textbf{70.2}\%  & 27.2\%  / \textbf{72.8}\%\\
        AViTAR w/o visual                                   & 39.6\% / \textbf{60.4}\%  & 46.3\%  / \textbf{53.9}\%\\
    \bottomrule
    \end{tabular}
    \vspace*{-0.1in}
    \caption{User study results. X\%/Y\% indicates among all paired examples for this baseline and AViTAR, X\% of participants prefer this baseline while \textbf{Y\%} prefer AViTAR.
    }
    \vspace{-0.15in}
    \label{tab:user_study}
\end{table}

Table~\ref{tab:user_study} \KG{shows the resulting preference scores}. Compared to each baseline, AViTAR is always preferred. Note that no participant has a background in acoustics, and some might simply pick the one that sounds \KG{``clean"} rather than having the correct room acoustics.  This may be the reason even the anechoic input has a higher preference score than the U-Net model. Despite the lack of domain knowledge, participants still consistently favor our model over other baselines. 

\vspace{-0.15in}
\paragraph{Qualitative examples.} 
Figure~\ref{fig:qual} \KG{shows example outputs.}
\KG{Please see the Supp.~video to gauge the audio quality.}

\vspace*{-0.05in}
\section{Conclusion}

\vspace*{-0.05in}
\KG{We proposed the visual acoustic matching task and introduced the first model to address it.  Given an image and audio clip, our method injects realistic room acoustics to match the target environment.  Our results validate their realism with both objective and perceptual measures. Importantly, the proposed model is trainable with unannotated, in-the-wild Web videos.  In future work we aim to extend our model to leverage the dynamics in target visual scenes in video.  \KGcr{We discuss potential societal impact in Supp.}}

\vspace{0.05in}

\noindent \textbf{Acknowledgements} \KGcr{UT Austin is supported in part by a gift from Google and the IFML NSF AI Institute.}
\clearpage

{\small
\bibliographystyle{ieee_fullname}
\bibliography{egbib}
}

\clearpage

\setcounter{section}{7}
\setcounter{table}{4}
\setcounter{figure}{5}

\section{Supplementary Material}
In this supplementary material, we provide additional details about:
\begin{enumerate}[itemsep=0mm]
    \item Supplementary video for qualitative assessment of our model's performance.
    \item Acoustic AVSpeech filtering process (\KG{referenced in Sec. 4 of the main paper}).
    \item Acoustic changes after each alteration step (\KG{referenced in} Sec. 5).
    \item Implementation and training details (\KG{referenced in} Sec. 6).
    \item Evaluation and baseline details (\KG{referenced in} Sec. 6).
    \item Ablations on acoustics alteration (\KG{referenced in} Sec. 6.2).
    \item \CAcr{Ablations on GAN losses (referenced in Sec. 6).}
    \item \CAcr{Sim2real generalization (referenced in Sec. 6).}
    \item \CAcr{Applicability on non-speech sounds (referenced in Sec. 4).}
    \item \CAcr{Interpretation of the neural network results.}
    \item \CAcr{Does the model capture room size?}
    \item User study interface. 
    \item Societal impact (\KG{referenced in} Sec. 7).
\end{enumerate}

\subsection{Supplementary Video}
This video includes examples generated by AViTAR and baselines for SoundSpaces-Speech and Acoustic AVSpeech. We also demonstrate application scenarios for augmented reality and video conferencing. Wear your headphones for a better listening experience.

\subsection{Acoustic AVSpeech Filtering Process}

\KG{As noted in the main paper, we apply a series of automatic filters to the AVSpeech dataset~\cite{ephrat2018looking} in order to select those clips relevant for our task.  Here we detail those steps.}

AVSpeech is a large-scale audio-visual dataset comprising speech video clips with no interfering background noises. The segments are 3-10 seconds long, and in each clip the audible sound in the soundtrack belongs to a single person speaking who is visible in the video. In total, the dataset contains roughly 4700 hours of video segments, from a total of 290k YouTube videos, spanning a wide variety of people, languages and face poses.

\KG{Since our dereverberation model used during acoustics alteration is} 
trained on an English corpus, we first run a language classification algorithm over all the \KG{AVSpeech} audio clips and remove clips where the spoken language is not English. After this step, there are still many videos which are almost anechoic, sometimes due to the audio being recorded post video recording, or to the speaker using a microphone very close to his/her mouth. To remove such examples, we train an RT60 predictor on the SoundSpaces-Speech (details in Sec.~\ref{sec:implementation}), run it on all AVSpeech clips and remove examples where the predicted RT60 is less than 0.1s. Lastly, we balance the distribution of RT60 such that it is not heavily skewed toward the anechoic side.

\subsection{Acoustic Changes After Each Alteration Step}
In Table~\ref{tab:acoustic_changes}, we show how the acoustics change after performing each step in the acoustics-alteration process by evaluating RT60 and MOS of the processed speech on the test split. 
\KG{What we expect to see is that the original audio gets cleaner via dereverberation, then becomes increasingly reverberant and noisy as we perform the subsequent steps that are designed to disguise the audio with other room acoustics from the sampled IR.  This is indeed what we observe.}
The original audio input has a high RT60 value on average, but after dereverberation the RT60 drastically goes down to 0.088s and the speech quality becomes better. After reverberating, the average RT60 goes up again, with a lower MOS score. Adding noise slightly improves the RT60 value and reduces the speech quality. For clean speech, its average RT60 is much lower and the MOS score is also high.   \KG{Note that here we show the MOS scores, not the MOS errors; higher values indicate higher quality speech.}

\begin{table}[]
    \centering
    \begin{tabular}{c|c|c}
    \toprule
    Acoustic Changes                        & RT60 (s) & MOS \\
    \midrule
        Original audio                      & 0.436 & 2.778 \\
        Dereverb.                           & 0.088 & 2.970 \\
        Dereverb. + Randomization           & 0.424 & 2.620 \\
        Dereverb. + Randomization + Noise   & 0.462 & 2.513 \\
        Clean                               & 0.049 & 3.285 \\
    \bottomrule
    \end{tabular}
    \vspace{-0.05in}
    \caption{Acoustic changes after each alteration step.}
    \label{tab:acoustic_changes}
\vspace{-0.2in}
\end{table}

\subsection{Implementation and Training Details}
The 1D convolutions for encoding and decoding the waveform have kernel sizes of $16,8,4,4$ and strides $8,4,2,2$ respectively. The total downsampling/upsampling rate $D$ is $128$. The latent feature size for $A_i$, $V_i$ and $M_i$ is $512$. The number of cross-modal encoders $N$ is $4$. There are 8 attention heads in each attention layer. The number of sub-discriminators $K$ is $3$ and $\lambda_1$ and $\lambda_2$ are $1$ and $45$, respectively. The learning rate for the generator and discriminators are $0.005$ and $0.002$. 

The input audio clip is 2.56 seconds for both datasets. On SoundSpaces-Speech, the input image size is $192 \times 576$, and we randomly shift the panoramic image during training for the model to learn viewpoint-invariant room acoustics features, following the original paper~\cite{chen_dereverb21}. On Acoustic Speech, the input image is first resized to $270 \times 480$, followed by random cropping to size $180 \times 320$ and random horizontal flip for data augmentation. We train all models 600 epochs on SoundSpaces-Speech and 300 epochs on Acoustic AVSpeech, and evaluate the checkpoint with the lowest validation loss on the test set. We will share the code and data upon acceptance.

\subsection{Evaluation and Baseline Details}\label{sec:implementation}
\paragraph{RT60 estimator.} On SoundSpaces-Speech, we have access to the reverberant speech clip as well as the impulse response. We first encode the 2.56s speech clips as spectrograms, process them with a ResNet18~\cite{resnet18} and predict the RT60 of the speech. The ground truth RT60 is calculated with the Schroeder method~\cite{Schroeder}. We optimize the MSE loss between the predicted RT60 and the ground truth RT60.

\vspace{-0.1in}
\paragraph{Image2Reverb~\cite{singh_image2reverb_2021}.} We obtained the code from the authors and made some changes to accommodate their model on our dataset. First of all, we replace the depth estimator with the ground truth depth image that we have access to on SoundSpaces-Speech. We also increase the size of the input image to match the size of the panorama. Lastly, we change the sampling rate from 22050 to 16000. The rest of the code stays the same, including the visual encoder pretrained on Places365 and the auxiliary loss on RT60 prediction.

\subsection{Ablations on Acoustics Alteration}
Table 3 shows ablations on the proposed acoustics-alteration strategy. Removing either the acoustic randomization or noise leads to worse generalization to novel sounds compared to the full process. This is because without these two steps, it is easier for the model to overfit the residual acoustic information in the dereverberated audio rather than use the visual content for recovering correct acoustics. If both are removed (``Dereverb.''), the model does not generalize to novel sounds. Similarly, the dereverberation step is also very important. If we simply randomize the acoustics with another IR and add noise to the original audio (``$A_T$ + Randomization + Noise``), there is no training sample that has less reverberation than the target audio, and the model simply learns to perform dereverberation; this leads to poor generalization as well. Altogether, all three steps are necessary to create acoustic mismatch with the image and force the model to recover the correct acoustics based on images.

\subsection{Ablations on GAN Losses}
\CAcr{Here we detail each GAN loss component and how they affect the performance.
Mel-spectrogram loss $\mathcal{L}_{Mel}$ is the L1 distance between two mel-spectrograms, which improves the perceptual quality. Feature matching loss $\mathcal{L}_{FM}$ is a learned similarity metric for features of the discriminator between two audio samples. We ablate these two loss terms separately and the results are shown in Table~\ref{tab:gan_loss_ablations}. Removing the Mel-spectrogram loss leads to a great reduction on all metrics. Removing the feature matching loss leads to higher STFT distance and RT60 error while marginally lower MOS error. Overall, these two ablations show both components are important for synthesizing realistic audio with matched acoustics.
}

\begin{table}[t]
\centering
\begin{tabular}{ c|c|c|c } 
 \toprule
       AViTAR                   & STFT & RTE (s)  & MOSE\\
\midrule
 Full model                     & \textbf{0.822} & \textbf{0.062} & 0.195 \\ 
 w/ $\mathcal{L}_{Mel}$         & 2.907 & 0.190 & 0.833\\
 w/ $\mathcal{L}_{FM}$          & 0.831 & 0.063 & \textbf{0.192}\\
 \bottomrule
\end{tabular}
    \vspace{-0.05in}
\caption{Ablations on GAN loss components. 
}
\vspace{-0.2in}
\label{tab:gan_loss_ablations}
\end{table}

\subsection{Sim2real Generalization}
\CAcr{
To understand how well the model trained on synthetic dataset generalizes to web videos, we train a new AViTAR model on SoundSpaces-Speech with only RGB input, and then test it on the Acoustic AVSpeech dataset, which yields RTE of 0.278s, MOSE of 0.898, while the model trained and tested on Acoustic AVSpeech gives 0.183s RTE and 0.453 MOSE (Table 1). The newly trained synthetic model tends to generate more reverberation, likely due to the visual discrepancy. This highlights the effectiveness of our self-supervised acoustic alteration strategy.
}

\subsection{Applicability on Non-speech Sounds.}
\CAcr{
To understand if our models applies to non-speech sounds, we train AViTAR on SoundSpaces by replacing the human speech with non-speech sounds, e.g. ringtone, music, etc., the model has 0.064 RTE on test-unseen, higher than human speech (0.062 RTE), while outperforming AV U-Net (0.074 RTE) and the input (0.176 RTE). So while we focus on speech for application reasons, this positive non-speech result makes sense because our model design is agnostic to the type of audio.}

\subsection{Interpretation of the Neural Network Results.}
\begin{figure}[t]
    \centering
    \includegraphics[width=0.8\linewidth]{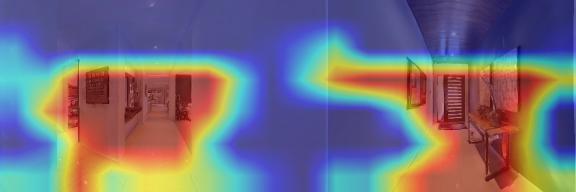}
    \vspace{-0.05in}
    \caption{Grad-CAM for corridor scene.}
    \label{fig:grad_cam}
\vspace{-0.1in}
\end{figure}

\begin{figure}[t]
    \centering
    \includegraphics[width=\linewidth]{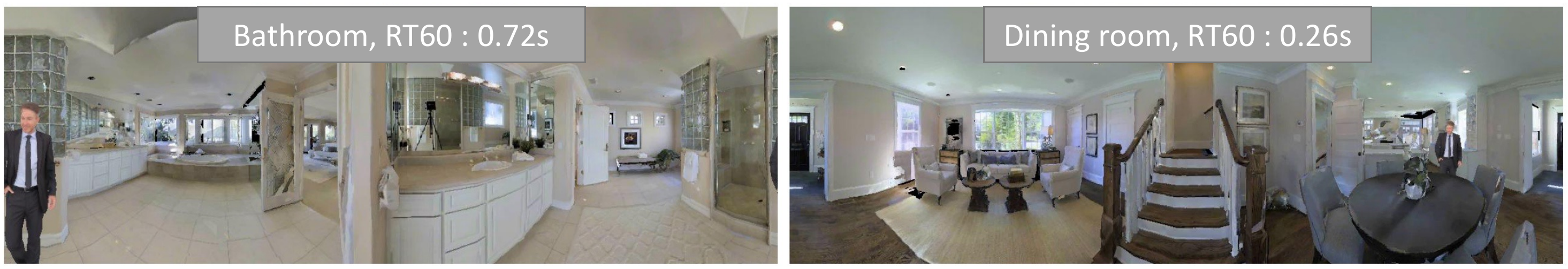}
    \vspace{-0.2in}
    \caption{Rooms of similar sizes but different acoustics.}
    \label{fig:room_comparison}
\vspace{-0.1in}
\end{figure}

\CAcr{
To show how the model understands the image, we can use Grad-CAM to visualize the activations. For example, in Fig.~\ref{fig:grad_cam} Grad-CAM highlights two sides of the corridor because they lead to longer reverberation. Fig.~\ref{fig:room_comparison} shows two rooms of similar sizes, and our model predicts longer RT60 for the bathroom likely because it has more reflective materials and leads to longer reverberation time.}

\subsection{Does the Model Capture Room Size?}
\begin{figure}[t]
    \centering
    \includegraphics[width=0.6\linewidth]{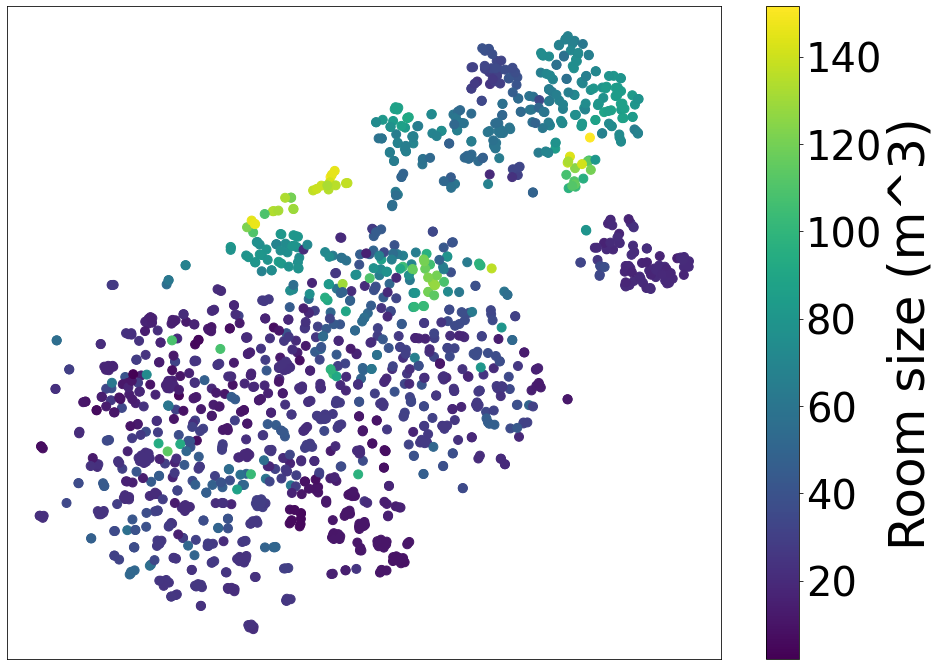}
    \vspace{-0.05in}
    \caption{T-SNE projection of visual features colored by room size}
    \label{fig:tsne_room_size}
\vspace{-0.1in}
\end{figure}

\CAcr{
To understand if our learned model captures room sizes, we check two things: 1) whether the learned visual features manage to pick up on room size (the clustered colors in Fig.~\ref{fig:tsne_room_size} suggest yes), and 2) whether we output only a narrow set of acoustics for the same room type (the distribution of RT60s over all kitchens in the test split (Table~\ref{tab:rt60_kitchen}) suggests no).
Furthermore, we project visual features on the 2D plane colored by visible room volume with T-SNE (shown in Figure~\ref{fig:tsne_room_size}). The gradient from small room volumes to large room volumes indicates that room size is captured in visual features. In addition, we show the distribution of RT60s over all kitchen environments in the test-unseen split in Table~\ref{tab:rt60_kitchen} and it is quite diverse. 
}

\begin{table}[t]
    \centering
    \setlength{\tabcolsep}{4pt}
    \scalebox{1}{
    \begin{tabular}{c|c|c|c|c|c}
    \toprule
    RT60 (s)        &  $\leq$ 0.2 & 0.2-0.3 & 0.3-0.4  & 0.4-0.5    & $\geq$ 0.5\\
    Percent (\%)    &  11.9       & 55.0    & 27.7     & 5.5        & 1.0 \\
    \bottomrule
    \end{tabular}}
    \vspace{-0.05in}
    \caption{RT60 distribution over kitchens \KG{(other scenes show similar diversity)}.}
    \label{tab:rt60_kitchen}
    \vspace{-0.1in}
\end{table}

\subsection{User Study Interface}
\begin{figure}[t]
    \centering
    \includegraphics[width=\linewidth]{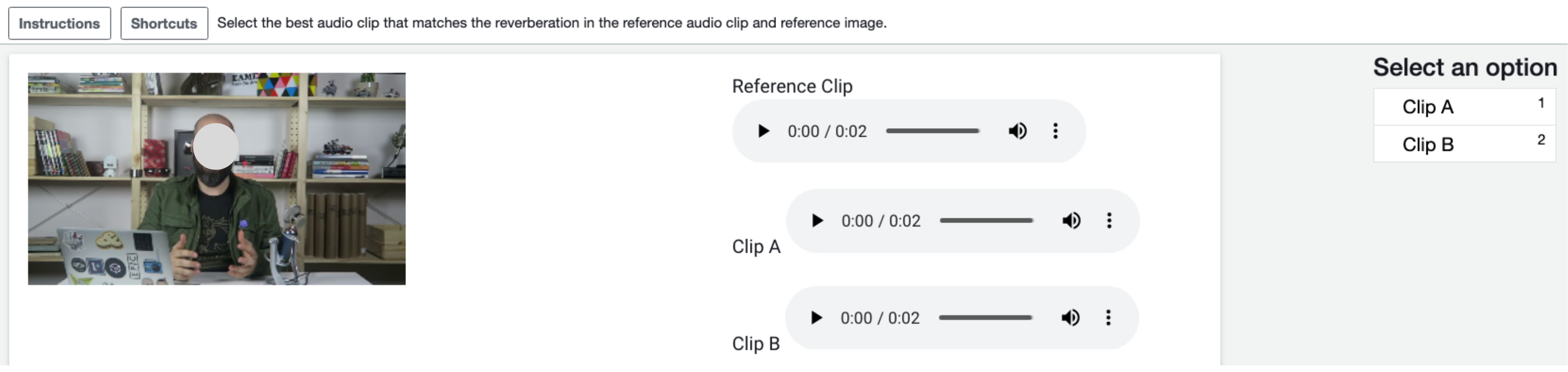}
    \caption{User study interface on MTurk. Given a reference image, a reference audio and two clips generated by AViTAR and a baseline (with shuffled order), participants are asked to pick the best clip that matches the reverberation in the reference image and audio.}
    \label{fig:mturk_interface}
        \vspace{-0.1in}
\end{figure}

Figure~\ref{fig:mturk_interface} shows the interface for our user study on MTurk. See details of the instruction in the caption.

\subsection{Societal Impact}
We believe this work can have a positive impact on many real-world applications, e.g., \KG{video editing}, film dubbing, and AR/VR, and discussed in the paper.
However, \KG{future applications built on such technology must also take care to avoid its misuse.  The ability to transform a voice to sound like it comes from a new environment could potentially} be misused for 
\KG{enhancing} deep fake videos, by matching an audio not recorded along with the video to the visual stream.

\end{document}